\documentclass[10pt,twocolumn,letterpaper]{article}

\usepackage[final]{cvpr}
\usepackage{graphicx}
\usepackage{amsmath}
\usepackage{amssymb}
\usepackage[title]{appendix}
\usepackage{xspace}
\usepackage{mathtools}
\usepackage{bbding}
\usepackage{amsmath}
\usepackage{amsthm}
\usepackage{booktabs}
\usepackage{verbatim}
\usepackage{multirow}
\usepackage{makecell}
\usepackage{algorithm}
\usepackage{algorithmic}
\usepackage{times}
\usepackage{epsfig}
\usepackage{pifont}
\usepackage{booktabs}
\usepackage{capt-of}
\usepackage{colortbl}
\usepackage{dsfont}
\usepackage{comment}
\usepackage[dvipsnames]{xcolor}
\usepackage{pifont}
\usepackage{bbm}
\usepackage{url}
\usepackage{nicefrac}
\usepackage{colortbl}
\usepackage{cases}
\definecolor{cvprblue}{rgb}{0.21,0.49,0.74}
\usepackage[pagebackref=true,breaklinks=true,citecolor=cvprblue,colorlinks,bookmarks=false]{hyperref}
\usepackage[capitalize]{cleveref}
\crefname{section}{Sec.}{Secs.}
\Crefname{section}{Section}{Sections}
\Crefname{table}{Table}{Tables}
\crefname{table}{Tab.}{Tabs.}

\newcommand{\lyx}[1]{\textcolor{black}{#1}}

\definecolor{baselinecolor}{gray}{.9}
\definecolor{darkgreen}{rgb}{0.13, 0.55, 0.13}

\newcommand{\tablestyle}[2]{\setlength{\tabcolsep}{#1}\renewcommand{\arraystretch}{#2}\centering\footnotesize}
\newcommand{\tablestylesmaller}[2]{\setlength{\tabcolsep}{#1}\renewcommand{\arraystretch}{#2}\centering\scriptsize}
\renewcommand{\paragraph}[1]{\vspace{1.25mm}\noindent\textbf{#1}}

\let\originalleft\left
\let\originalright\right
\renewcommand{\left}{\mathopen{}\mathclose\bgroup\originalleft}
\renewcommand{\right}{\aftergroup\egroup\originalright}

\usepackage{caption}
\usepackage{fancyhdr}

\begin{document}
\title{Memory Consistency Guided Divide-and-Conquer Learning \\ for Generalized Category Discovery}


\author{Yuanpeng Tu$^{1}$ \quad Zhun Zhong$^{2}$ \quad Yuxi Li$^{3}$ \quad Hengshuang Zhao$^{1\dagger}$\\
$^{1}$The University of Hong Kong ~~~~~~~~ $^{2}$University of Nottingham ~~~~~~~~ $^{3}$Tencent \\ 
}

\maketitle

\begin{abstract}
    Generalized category discovery (GCD) \lyx{aims at addressing} a more realistic and challenging setting of semi-supervised learning, where only part of the category \lyx{labels are assigned to certain training samples}. Previous methods generally employ \lyx{naive} contrastive learning or unsupervised clustering scheme for all the samples. \lyx{Nevertheless, they usually ignore} the inherent critical information within the historical predictions of the model \lyx{being trained}. Specifically, we empirically reveal that a significant number of \lyx{salient} unlabeled samples \lyx{yield} consistent historical predictions corresponding to their ground truth category. From this observation, we propose a \textbf{M}emory \textbf{C}onsistency guided \textbf{D}ivide-and-conquer \textbf{L}earning framework (MCDL). \lyx{In this framework, we introduce two memory banks to record historical prediction of unlabeled data, which are exploited to measure the credibility of each sample in terms of its prediction consistency.  With the guidance of credibility, we can design} a divide-and-conquer learning strategy to fully utilize the discriminative information \lyx{of unlabeled data while alleviating} the negative influence of noisy labels. Extensive experimental results on multiple benchmarks demonstrate the generality and superiority of our method, where our method outperforms state-of-the-art models by a large margin on both seen and unseen classes of the generic image recognition and challenging semantic shift settings (i.e., with +8.4\% gain on CUB and +8.1\% on Standford Cars). 

\end{abstract}
\renewcommand{\thefootnote}{}
\footnotetext{$\dagger$ Corresponding author.}

\section{Introduction}
\thispagestyle{plain}

\label{sec:intro}
While current methods outperform humans in recognizing images of diverse styles using large-scale labeled benchmarks, such extensive human annotations are not always available for training. Therefore, a large number of recognition models focus on extracting information from abundant unlabeled data. Semi-supervised learning (SSL)~\cite{sohn2020fixmatch, berthelot2019mixmatch} is a well-known and promising \lyx{learning paradigm to fulfill this goal}, which aims to achieve nearly the same performance as supervised \lyx{counterparts} with only a few annotations for training. However, SSL assumes that labeled instances are available for all categories that are expected to classify during inference. To \lyx{extend SSL into a more general scenario}, generalized category discovery (GCD)~\cite{vaze2022generalized} is recently proposed, which considers that the \lyx{semantic categories of} unlabeled data \lyx{are a superset of categories in labeled ones.} The goal of GCD is to accurately classify already-seen classes in the labeled data as well as recognize unseen novel classes in the unlabeled data. A well-developed GCD method can help extend existing category taxonomy and significantly reduce expensive labeling cost.

\begin{figure}[!t]
    \centering
    \includegraphics[width=1.0\linewidth]{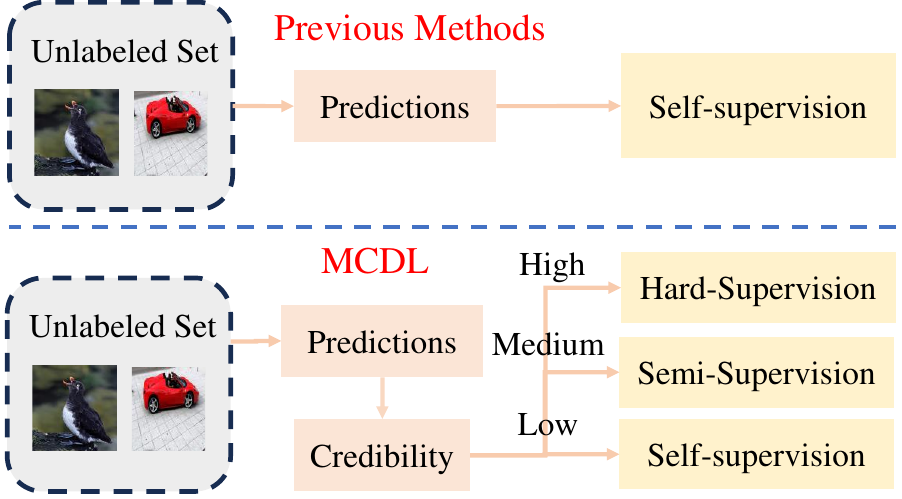}
    \caption{\textbf{Comparison between previous methods and our proposed MCDL.} MCDL is trained in a divide-and-conquer manner based on the credibility learnt from historical predictions.
    }
    \label{fig:intro}
\end{figure}

\begin{figure*}[!t]
    \centering
    \includegraphics[width=1.0\linewidth]{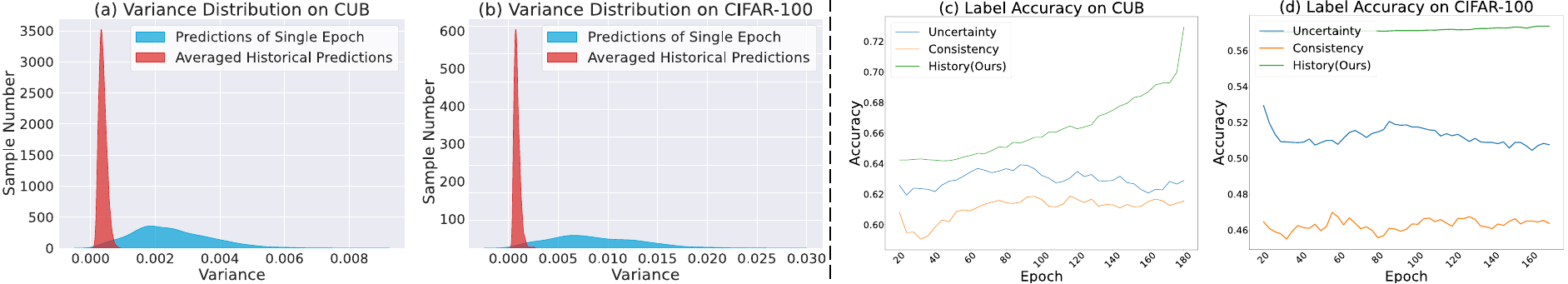}
    \caption{\textbf{The variance distribution of predictions} of the 20th epoch and averaged historical predictions of 10-20 epochs on the CUB (a) and CIFAR-100 (b) datasets respectively. \textbf{Label accuracy} of uncertainty-based/consistency-based and our history-based method (MCDL) for the selected samples on the CUB (c) and CIFAR-100 (d) datasets. Consistency based method selects samples that have the consistent predictions between the weakly-augmented and strong-augmented views, while the uncertainty based one chooses samples that have the top-10 samples with the smallest uncertainty in each category. 
    }
    \label{fig:accuracy}
\end{figure*}

Previous investigations~\cite{vaze2022generalized,cao2021open,fini2021unified,han2021autonovel} mainly solve the GCD task from two perspectives: the first involves \lyx{learning} universal feature representations to expedite the discovery of novel categories, while the second entails generating pseudo cluster labels for unlabeled data to guide training. The former \lyx{usually resorts to} self-supervised learning techniques~\cite{caron2021emerging,gidaris2018unsupervised,
he2020momentum,zhao2020distilling}, which enhance representation generalization towards unfamiliar categories. \lyx{As for the latter ones,} earlier research ~\cite{zhao2021novel,han2021autonovel,zhong2021openmix} has embraced parametric approaches, constructing a trainable classifier based on extracted features and concurrently optimizing the underlying backbone using both labeled and pseudo-labeled set.

However, as illustrated in Fig.~\ref{fig:intro}, conventional methods treat all unlabeled samples equally by using an unsupervised/self-supervised scheme. These approaches largely ignore the \lyx{various credibility} inherent in the predictions of unlabeled samples. Previous SSL works~\cite{sohn2020fixmatch,berthelot2019mixmatch} reveal that samples with consistent outputs \lyx{across different views are more trustworthy in their predictions}. However, due to the unstable training process, \lyx{it is unreliable to select} samples based on prediction \lyx{consistency} from a single \lyx{epoch}. As shown in Fig.~\ref{fig:accuracy}(a) and (b), predictions exhibit significant variance across different \lyx{epochs} in both generic and fine-grained benchmarks. Therefore, selecting samples based on uncertainty~\cite{wang2021combating,wang2020double} or cross-view consistency~\cite{zhang2021flexmatch,sohn2020fixmatch} \lyx{in a single epoch} only achieves inferior \lyx{quality of pseudo labels}, as shown in Fig.~\ref{fig:accuracy} (c) and (d). 
Furthermore, training with such selected samples fails to clearly improve the pseudo label accuracy throughout the process, especially in the middle and later stages.
In contrast, when selecting samples based on historical predictions \lyx{across multiple epochs}, the results exhibit more reliable statistics with smaller variances and a significant improvement in the \lyx{pseudo} label quality is observed \lyx{if training samples are} selected based on the consistency of historical predictions.

Motivated by this observation, we \lyx{propose a Memory Consistency-guided Divide-and-conquer Learning paradigm (MCDL), which} focuses on performing adaptive sample credibility modeling based on historical predictions. Specifically, we \lyx{design a Dual Consistency Modeling strategy (DCM), which} constructs two online-updating memory banks to maintain the historical predictions from weakly-augmented and strongly-augmented views of each sample. Subsequently, the credibility of each sample is modeled based on intra-memory and inter-memory consistency. With the learned credibility, we design a divide-and-conquer strategy to fully harness the discriminative information present in samples of different credibility levels. Samples with high, medium, and low credibility are utilized for supervised, semi-supervised, and self-supervised learning respectively. \lyx{This adaptive strategy effectively helps} mitigate the negative influence of noise in the generated pseudo labels and thus consistently improves the accuracy of all classes. The contributions of our proposed MCDL can be summarized as follows:
\begin{itemize}
\item We present a simple but effective framework for GCD, namely MCDL, which performs Divide-and-Conquer Learning (DCL) for different unlabeled samples based on their \lyx{various data credibility}.

\item \lyx{In MCDL, we propose Dual Consistency Modeling strategy (DCM) to precisely measure the credibility of different data}. Predictions from two types of augmentations are utilized to construct complementary online-updating memory banks, which can model the sample credibility based on intra/inter-memory consistency. 

\item MCDL can achieve state-of-the-art performance and exhibit significant superiority to previous approaches across both various generic and challenging GCD benchmarks. In addition, it can be easily integrated with existing methods as a plug-in-play module to boost their performance.

\end{itemize}

\section{Related work}

\begin{figure*}[!t]
    \centering
    \includegraphics[width=1.0\linewidth]{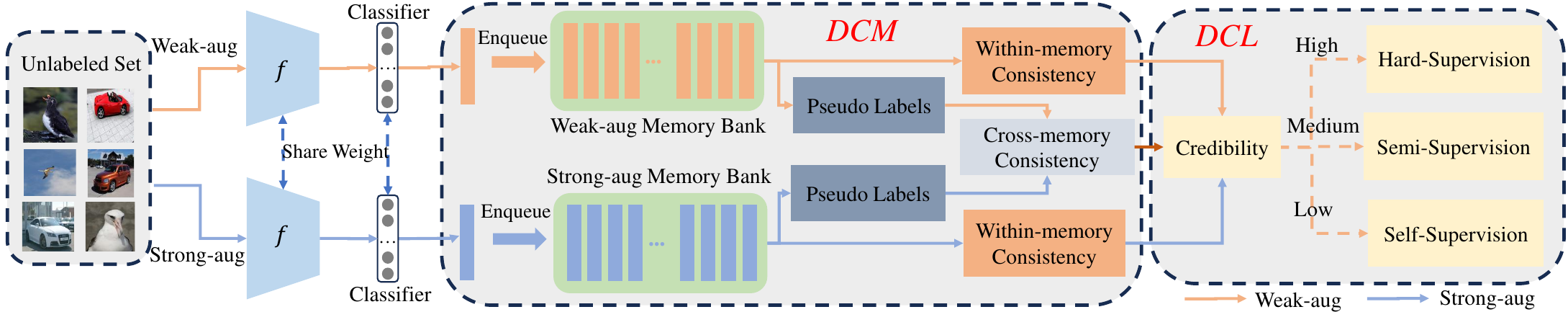}
    \caption{\textbf{An overview of the proposed MCDL method.} (a) DCM: Dual-consistency Credibility Modeling~(Sec. \ref{sec:32}). (b) DCL: Divide-and-Conquer Learning~(Sec. \ref{sec:33}). The samples are first adaptively assigned with credibility based on their historical predictions and then tackled with three different learning strategies based on their credibility-levels in a divide-and-conquer manner.
    }
    \label{fig:framework}
\end{figure*}

\noindent\textbf{Generalized category discovery.} Novel category discovery (NCD) is first formalized as cross-task transfer in \cite{han2019learning}, which targets at discovering unseen categories from unlabeled data that has non-overlapped classes with the labeled ones. Earlier works~\cite{hsu2017learning} mostly maintain two networks for learning from labeled and unlabeled data respectively. ~\cite{han2021autonovel} introduces a three-stage framework. Specifically, the model is firstly trained with the whole dataset in a self-supervised manner and then fine-tuned only with the fully-supervised labeled set to capture the semantic knowledge for the final joint-learning stage. ~\cite{zhong2021openmix} tackles the problem by mixing the labeled and unlabeled set to prevent model from over-fitting to labeled categories. Similarly, \cite{zhong2021neighborhood} generates pairwise pseudo labels for unlabeled data and mixes samples in the feature space to construct hard negative pairs. 

To address this issue, Generalized Category Discovery (GCD) is proposed, which does not have the assumption in the NCD and is first comprehensively formalized in \cite{vaze2022generalized}. It constructs a simple framework to discover unseen categories by combining contrastive learning and semi-supervised learning. For example, \cite{lin2020discovering} addresses this issue by generating pseudo labels with unsupervised clustering. \cite{wen2023simgcd} finds that performance degradation of previous methods is mainly due to the misuse of features and unreliable pseudo labelling strategy and propose a simple parametric classification method to tackle with the problem. Afterward, CiPR~\cite{hao2023cipr} utilizes the cross-instance positive relations to construct positive and negative pairs for contrastive learning and introduce a hierarchical clustering method to boost representation quality. Similarly, PromptCAL~\cite{zhang2023promptcal} focuses on utilizing reliable pairwise sample affinities for better semantic clustering of class tokens and visual prompts based on online-updating iterative semi-supervised affinity graphs. Moreover, GPC~\cite{zhao2023learning} discovers that class number estimation and representation learning can compensate for each other and propose a EM-like framework by introducing Gaussian Mixure Model (GMM) and prototype-based contrastive learning. In \cite{yang2022divide}, it proposes a compositional expert network to tackle new and old classes with different networks and performs global-to-local alignment/aggregation to reinforce the quality of pseudo labels. However, all these methods fail to harness the discriminative information existed in the historical predictions, resulting in poor performance on the unseen categories.

\noindent\textbf{Semi-supervised learning} (SSL) is a critical area of research with a range of proposed methods~\cite{berthelot2019mixmatch, zhai2019s4l,sohn2020fixmatch}. In SSL, the assumption is that labeled instances cover all possible categories \lyx{presented} in the unlabeled dataset. The ultimate objective is to develop a model capable of classification by leveraging both labeled samples and the abundant unlabeled data available. A particularly effective approach in SSL is the consistency-based methodology, which obliges the model to learn cohesive representations from two distinct augmentations of a single image~\cite{berthelot2019mixmatch, tarvainen2017mean, sohn2020fixmatch}. Furthermore, the effectiveness of self-supervised representation learning has emerged, demonstrating its utility in SSL~\cite{zhai2019s4l, rebuffi2020semi}. This self-supervised approach produces robust representations that greatly benefit the SSL task.

\section{Methodology}


\noindent\textbf{Problem statement}. Following settings in ~\cite{vaze2022generalized}, the training dataset is denoted as $\mathcal{D}=\mathcal{D}_l \cup \mathcal{D}_u$, where the labeled set is $\mathcal{D}_l=\left\{\boldsymbol{x}_i, y_i\right\}_{i=1}^{N_l} \subset \mathcal{X}_l \times \mathcal{Y}_l$ and the unlabeled set is denoted as $\mathcal{D}_{u}=\{x_i^u\}_{i=1}^{N_u} \subset \mathcal{X}_u$. $N_l$ and $N_u$ indicate the number of labeled and unlabeled samples respectively.
The label space of the $\mathcal{D}_l$ only include already-seen classes: $\mathcal{Y}_l=\mathcal{C}_{old}$, while the $\mathcal{D}_u$ comprises both already-seen and new classes: $\mathcal{Y}_u= \mathcal{C} = \mathcal{C}_{old} \cup \mathcal{C}_{new}$, where $\mathcal{C}$ denotes all the categories of $\mathcal{D}$ and is known or can be obtained with off-the-shelf approaches. The goal of GCD is accurately recognizing both the already-seen and new classes.

\noindent\textbf{Overview}. The overview of our MCDL is shown in Fig.~\ref{fig:framework}. The model architecture consists of a encoder network $f(\cdot)$ and a classifier denoted as $g(\cdot)$.
Our MCDL comprises of two main modules: dual-consistency credibility modeling (DCM) and divide-and-conquer learning (DCL). The predictions of the weakly-augmented and strongly-augmented\footnote{For all data-related symbols, we use the subscript of $w$/$s$ to denote the view from weak/strong augmentation.} unlabeled samples 
are first fed into the DCM to perform adaptive credibility modeling based on the consistency of historical predictions. Subsequently, the samples are divided into three categories based on their credibility levels (high, medium, and low). Each category is tackled with different \lyx{learning strategies} (hard, semi, and self-supervision) in the DCL respectively to effectively leverage the inherent supervisory signals. 

\subsection{Dual-consistency credibility modeling}
\label{sec:32}
As shown in Fig.~\ref{fig:intro}, samples with consistently predicted categories are more likely to represent their true classes. Therefore, we create online-updating memory banks to store predictions for unlabeled samples. Since weak augmentation does not significantly alter predictions, different historical epochs may yield similar results in the memory bank for each weak-augmented sample. \lyx{However, the intrinsic} confirmation bias can cause the model to increasingly over-fit to noisy pseudo labels derived from these memory banks, resulting in inaccurate sample selection \lyx{and hindering model training}. In order to mitigate the impact of confirmation bias and introduce \lyx{proper} disturbance during model training, we also construct a memory bank using predictions from strongly-augmented images. By combining the information from both memory banks, we can achieve more accurate sample selection.

\noindent\textbf{Memory bank construction}. Specifically, we denote the weakly-augmented/strongly-augmented batch of all the unlabeled samples as $\{x_i^u\}_{i=1}^{N_u}$. The memory banks of $x_i^u$ constructed by predictions of its weakly-augmented/strongly-augmented views are denoted $\mathcal{M}_w^i$ and $\mathcal{M}_s^i$ respectively. The $\mathcal{M}_w^i$ and $\mathcal{M}_s^i$ of $m$-th epoch are formulated as follows:
\begin{equation}
\begin{aligned}
& \mathcal{M}_w^i=\left[\begin{array}{llll}
p_w^{m-\mu+1}, & p_w^{m-\mu+2}, & \ldots, & p_w^m
\end{array}\right]^i_m, \\
& \mathcal{M}_s^i=\left[\begin{array}{llll}
p_s^{m-\mu+1}, & p_s^{m-\mu+2}, & \ldots, & p_s^m
\end{array}\right]^i_{m},
\end{aligned}
\end{equation}
where $\mu$ is the length of memory bank and \lyx{$p^m \in (0, 1)^{|\mathcal{C}|}$ is the probabilistic prediction output by classifier at epoch $m$}. Then we record all the predicted categories within $\mathcal{M}_w^i$ and $\mathcal{M}_s^i$ and calculate the occurrence count of each category in the memory bank:
\begin{equation}
\mathbb{M}_w^i=\operatorname{count}\left(\left[\operatorname{argm}\left(p_w^{m-\mu+1}\right), \ldots, \operatorname{argm}\left(p_w^m\right)\right]^i_m\right), \\
\end{equation}
\vspace{-3mm}
\begin{equation}
\mathbb{M}_s^i=\operatorname{count}\left(\left[\operatorname{argm}\left(p_w^{m-\mu+1}\right), \ldots, \operatorname{argm}\left(p_w^m\right)\right]^i_m\right),
\end{equation}
where $\operatorname{count}$ denotes calculating the occurrence count of different prediction categories and $\operatorname{argm}$ denotes $\operatorname{argmax}$.

\noindent\textbf{Credibility evaluation}. Afterward, we record the category with the highest number in $\mathbb{M}_w^i$ and $\mathbb{M}_s^i$ respectively, which are utilized as the metric of historical consistency. For more accurate credibility modeling, we take both intra-memory and inter-memory consistency into consideration. Specifically, only when there is high historical consistency in both $\mathbb{M}_w^i$ and $\mathbb{M}_s^i$ and the category with the highest number in $\mathbb{M}_w^i$ and $\mathbb{M}_s^i$ is the same one, the sample can be categorized as \lyx{data of high-credibility}. The process can be formulated by the conditions below:


\begin{subequations}
\label{eq:condition}
  \begin{numcases}{}
      \max \left(\mathbb{M}_w^i\right)>\mu*3/4, \label{eq:condition1} \\
      \max\left(\mathbb{M}_w^i\right)>\mu/4, \label{eq:condition2} \\
\operatorname{argm}\left(\mathbb{M}_w^i\right)=\operatorname{argm}\left(\mathbb{M}_s^i\right). \label{eq:condition3}
  \end{numcases}
\end{subequations}
In this paper, samples satisfying all the conditions in Eq.~\ref{eq:condition} are assigned with high credibility. Moreover, samples with slightly lower consistency still contain valuable discriminatory information in their historical predictions. Consequently, we proceed to select samples with medium credibility from the remaining samples. Specifically, when the sample satisfies only Eq.~\ref{eq:condition1} and ~\ref{eq:condition2}, it will be assigned with medium credibility. For the remaining samples, as they lack consistency and provide limited supervisory signals from their predictions, we assign them with low credibility. Here, we denote the sample set with high/medium/low-credibility as $\mathcal{D}_u^{high}/\mathcal{D}_u^{mid}/\mathcal{D}_u^{low}$ respectively for convenience.

\begin{figure}[!t]
    \centering
    \includegraphics[width=1.0\linewidth]{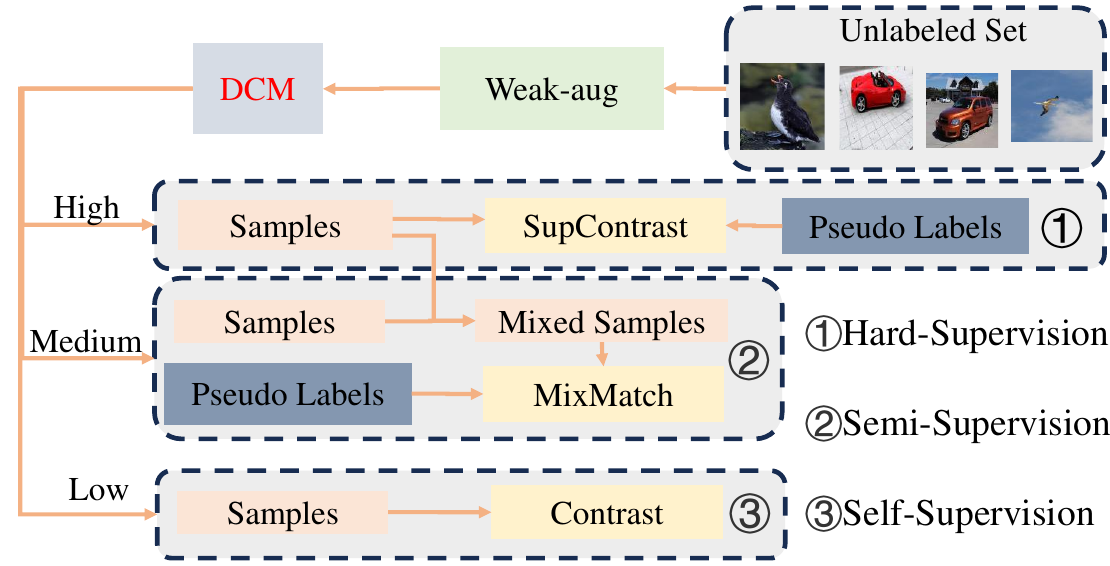}
    \caption{\textbf{The proposed divide-and-conquer learning strategy}, where samples with three types of credibility are tackled with different schemes respectively.
    }
    \label{fig:dclpip}
\end{figure}

\subsection{Divide-and-conquer learning}
\label{sec:33}
As shown in Fig.~\ref{fig:dclpip}, to fully harness the discriminative information within $\mathcal{M}_w^i$ and $\mathcal{M}_s^i$ and meanwhile alleviate the negative influence of noisy pseudo labels, we divide the unlabeled samples into three categories based on the credibility modeled by DCM and perform divide-and-conquer learning for each category.

\paragraph{High-credibility samples} \lyx{are more reliable and usually yields pseudo labels identical to their} ground truth class, so we directly use it for supervised learning together with the labeled set. Specifically, the supervised contrastive loss is used and the supervised labels $y_i^{high}=\{\operatorname{argm}(\mathbb{M}_w^i)\}$ are utilized for constructing positive pairs. The loss of $i$-th batch sample is formulated as:
\begin{equation}
\mathcal{L}_{ {sup}}= \frac{1}{\left|\mathcal{D}_u^{high}\right|} \sum_{i=1}^{\left|\mathcal{D}_u^{high}\right|} \frac{1}{\left|\mathcal{N}_i\right|} \sum_{q \in \mathcal{N}_i}-\log \frac{\exp \left(\boldsymbol{z}_i^{\top} \boldsymbol{z}_q / \tau_s\right)}{\sum_{n \neq i} \exp \left(\boldsymbol{z}_i^{\top} \boldsymbol{z}_n / \tau_s\right)},
\label{eq:sup}
\end{equation}
where $z$ denotes the feature from the encoder network: $z=f(x)$. And the $\tau_s$ denotes the scalar temperature parameter and the $\mathcal{N}_i$ indexes all other images \lyx{sharing the same label as $x_i$ in the batch.} 

\paragraph{Medium-credibility samples} contain valuable supervisory signals in their predictions as well but there exists a higher noise rate in the pseudo labels generated by their memory banks. Therefore, we utilize these samples in a semi-supervised manner and assist the training with $\mathcal{D}_u^{high}$ to alleviate the negative influence of noisy labels. The $\mathcal{D}_u^{high}$ is utilized as the labeled set and the $\mathcal{D}_u^{mid}$ is regarded as the unlabeled set. And the mean historical prediction of $\mathcal{M}_w^i$ and $\mathcal{M}_s^i$ is used as the pseudo labels of the unlabeled set, which can be denoted as:
\begin{equation}
    y_{i}^{mid} = (\overline{\mathcal{M}_w^i} + \overline{\mathcal{M}_s^i})/(2*\tau_u),
\end{equation}
where $\overline{\mathcal{M}}$ denotes the mean prediction of the corresponding memory bank and $\tau_u$ denotes the temperature scale parameter. Then we follow MixMatch~\cite{berthelot2019mixmatch} as the semi-supervised scheme to mix the data, where each sample is interpolated with another sample randomly selected for the sample set $\mathcal{D}_u^{high} \cup \mathcal{D}_u^{mid}$. Specifically, for a pair of samples $(x_1,x_2)$ with their corresponding labels $(y_1,y_2)$, the mixed sample and label $(\tilde{x},\tilde{y})$ is calculated by:

\begin{equation}
\begin{aligned}
\delta & \sim \operatorname{Beta}(\alpha, \alpha), ~~\delta^{\prime} =\max (\delta, 1-\delta), \\
\tilde{x} & =\delta^{\prime} x_1+\left(1-\delta^{\prime}\right) x_2,~~ \tilde{y} =\delta^{\prime} y_1+\left(1-\delta^{\prime}\right) y_2.
\end{aligned}
\end{equation}

\noindent $\alpha$ is a hyper-parameter and we conform to ~\cite{berthelot2019mixmatch} to set $\alpha=0.5$. \noindent MixMatch transforms $\mathcal{D}_u^{high}$ and $\mathcal{D}_u^{mid}$ into $\mathcal{\widetilde{D}}_u^{high}$ and $\mathcal{\widetilde{D}}_u^{mid}$ respectively. The loss used for $\mathcal{\widetilde{D}}_u^{high}$ is conventional cross-entropy loss, while the loss for $\mathcal{\widetilde{D}}_u^{mid}$ is mean squared loss. Thus the loss function can be formulated as:
\begin{equation}
\begin{split}
\mathcal{L}_{{semi}} =-\frac{1}{\left| \mathcal{\widetilde{D}}_u^{high} \right|} \sum_{x, y \in \mathcal{\widetilde{D}}_u^{high}} \sum_c y_c \log \left(g(f(x))\right) + \\
\frac{1}{\left|\mathcal{\widetilde{D}}_u^{mid}\right|} \sum_{x, y \in \mathcal{\widetilde{D}}_u^{mid}}\left\|y-g(f(x))\right\|_2^2 .
\end{split}
\label{eq:semi}
\end{equation}
With the assistance of $\mathcal{D}_u^{high}$, the negative influence of noise existed in the pseudo labels of $\mathcal{D}_u^{mid}$ can be effectively alleviated. Meanwhile the valuable supervisory information in the historical predictions can be utilized to boost the performance on unseen categories.

\paragraph{Low-credibility samples} include little discriminative information in their predictions. Thus we utilize them for training with a self-supervised loss. In addition, this loss is also applied for $\mathcal{D}_u^{high}$ and $\mathcal{D}_u^{mid}$ to enhance consistency of predictions. Specifically, the self-supervised loss is denoted as:

\begin{equation}
\mathcal{L}_{{self}} =\frac{1}{\left|\mathcal{D}_u\right|} \sum_{i=1}^{\left|\mathcal{D}_u\right|} \sum_c-{q_c^i}/\tau_{u} \log \left(p_c^i\right),
\label{eq:low}
\end{equation}
where $q$ denotes the prediction result of samples from another augmented view. 

\noindent\textbf{Overall loss}. The complete loss function of our proposed MCDL can be formulated as:

\begin{equation}
\label{eq:all}
\mathcal{L} = \mathcal{L}_{{sup}} + \lambda(\mathcal{L}_{{semi}}+\mathcal{L}_{{self}}),
\end{equation}
where $\lambda$ is a balance parameter for three losses. Putting this all together, Algorithm \ref{pseudoalgorithm} delineates the proposed MCDL for unlabeled data. Firstly, batch samples are divided into three categories based on their credibility modeled by DCM. Then samples with high/medium/low credibility are utilized for fully-supervised/semi-supervised/self-supervised learning respectively in DCL to harness the inherent semantic information and alleviate the negative influence of noisy pseudo labels. Finally, the predictions of current batch are utilized to update the online-updating memory banks. For labeled data, we adopt the same supervised contrastive learning loss as for high-credibility  unlabeled samples (Eq.~\ref{eq:sup}). We use it along with the MCDL loss (Eq.~\ref{eq:all}) but omit the description of this loss for simplicity.

\begin{algorithm}[!t] 
    \footnotesize
    \caption{The proposed MCDL framework}
    \label{alg:dmlp}
    \begin{algorithmic}[1]
    \renewcommand{\algorithmicrequire}{\textbf{Input:}}
    \REQUIRE Unlabeled set $\mathcal{D}_u$, encoder $f(\cdot)$, classifier $g(\cdot)$, batch size $b$, max iterations $m$. 
    \renewcommand{\algorithmicrequire}{ \textbf{Procedure:}}
    \REQUIRE
    \FOR{$i = 1$ to $m$}
    \STATE $\{x^i_u\}_w$ $\{x^i_u\}_s$ $\leftarrow$ SampleMiniBatch($D_u, b$).
    \STATE Feed $\{x^i_u\}_w$,$\{x^i_u\}_s$ into $f(\cdot)$ and $g(\cdot)$ and obtain predictions $\{p^i_l\}_w$, $\{p^i_u\}_s$.
    \STATE Update $\mathcal{M}_w, \mathcal{M}_s$ with $\{p^i_l\}_w$, $\{p^i_u\}_s$.
    \STATE Calculate credibility based on Eq.~\ref{eq:condition}. 
    \STATE Divide $\{x^i_u\}_w$ into $\mathcal{D}_u^{high}, \mathcal{D}_u^{mid}, \mathcal{D}_u^{low}$ based on the credibility.
    \STATE Feed predictions of $\mathcal{D}_u^{high}$ to Eq.~\ref{eq:sup}.
    \STATE Feed $\mathcal{D}_u^{high}, \mathcal{D}_u^{mid}$ to Eq.~\ref{eq:semi}.
    \STATE Feed $\{p^i_l\}_w$ to Eq.~\ref{eq:low}.
    \STATE Update parameters of $f(\cdot)$, $g(\cdot)$ in backward process.
    \ENDFOR
    \renewcommand{\algorithmicensure}{\textbf{Output:}}
    \ENSURE The final encoder $f(\cdot)$, classifier $g(\cdot)$.
    \end{algorithmic}  
    \label{pseudoalgorithm}
     
\end{algorithm}

\section{Experiments}

\subsection{Experimental setup}

\textbf{Datasets.} We substantiate the effectiveness of our MCDL through comprehensive validation on six distinct image recognition benchmarks, which encompass both generic object recognition datasets (namely CIFAR-10/100~\cite{krizhevsky2009learning}, ImageNet-100~\cite{tian2020contrastive}) and semantic shift benchmark suites (SSB), where four fine-grained datasets CUB~\cite{welinder2010caltech}, Standford Cars~\cite{krause20133d}, Herbarium19~\cite{herbarium-2020-fgvc7} and Aircraft~\cite{maji2013fine} are involved. Since SSB introduces fine-grained categories and data imbalance, it presents an additional challenge to the performance and robustness of the methods in real-world scenarios. For fair comparison, we adhere to the widely-used GCD setting~\cite{vaze2022generalized} to partition each dataset into labeled and unlabeled subsets. Concretely, half of the images belonging to the $\mathcal{C}_{old}$ known categories are randomly sampled from the dataset to construct the $\mathcal{D}_l$, and the remaining ones are adopted as the unlabeled subset $\mathcal{D}_u$, which contains all the classes $\mathcal{C}$ in the original dataset. Detailed statistics of the split protocol are shown in Tab.~\ref{tab:protocol}.

\begin{table}
\centering
\resizebox{\linewidth}{!}{ 
\begin{tabular}{lccccccc}
\toprule
     & CIFAR10 &\makecell{CIFAR \\ 100}  & \makecell{ImageNet \\ 100} & CUB  & \makecell{Standford \\ Cars}  &\makecell{Herbarium \\ 19}   & \makecell{Air \\ Craft}\\ 
\midrule
$|\mathcal{Y_L}|$ & 5       & 80       & 50           & 100  & 98    & 341  & 50 \\ 
$|\mathcal{Y_U}|$ & 10      & 100      & 100          & 200  & 196   & 683  & 50  \\ 
\midrule
$|\mathcal{D_L}|$ & 12.5k   & 20k      & 31.9k        & 1.5k & 2.0k  & 8.9k & 1.7k \\ 
$|\mathcal{D_U}|$ & 37.5k   & 30k      & 95.3k        & 4.5k & 6.1k  & 25.4k & 5.0k \\ 
\bottomrule
\end{tabular}
}
\caption{
%
\textbf{Split protocols} for the seen and unseen classes of the datasets evaluated in our experiments.
}
\label{tab:protocol}
\vspace{-0.6em}
\end{table}

\paragraph{Evaluation protocol.} In line with previous practices, we evaluate the model's performance using clustering accuracy (ACC) determined by the Bipartite Graph Matching. Specifically, during evaluation, we compare the predicted labels $\hat{y}$ to the ground truth label set $y^*$. The ACC is calculated as ACC=$\frac{1}{M}\sum_{i=1}^{M} \mathbb{E}(p(i)=\hat{y}(i))$, where $M=|D^u|$ and $p$ represents the optimal permutation that matches the predicted cluster assignments to the ground truth class labels, $\mathbb{E}$ outputs 1 when $p(i)=\hat{y}(i)$ otherwise 0. Additionally, we acknowledge the inherent imbalance in the division utilized in Herbarium 19~\cite{herbarium-2020-fgvc7}. Consequently, we report a balanced ACC, computed by averaging the ACC values on a per-class basis. This approach aims to mitigate potential biases resulting from the imbalanced evaluation.

\paragraph{Implementation details}. In accordance with previous GCD works, we utilize a ViT-B/16 network pre-trained with DINO as the backbone. During the training process, we freeze all the blocks except for the last block of the backbone. And we use RandAugment~\cite{cubuk2020randaugment} as the strong augmentation. $\tau_s$ and $\tau_u$ are set as 0.04 and 0.7 respectively. The batch size is set to 128, and the learning rate is initially set to 0.1, decaying according to a cosine schedule. Consistent with previous approaches, we train all the methods for 200 epochs on each dataset and report the accuracy of the models that achieve the highest performance on the validation set of the labeled categories. Furthermore, we set the balancing factor $\lambda$ to 0.35, and all experiments are conducted using an NVIDIA GeForce RTX 3090 GPU. In the experiments, we combine MCDL with the popular SimGCD~\cite{wen2023simgcd}.

\subsection{Experimental results}

\begin{table}[t]
\begin{center}
\tablestylesmaller{3.4pt}{1.05}
\begin{tabular}{lccccccccc}
\toprule
{\multirow{2}{*}{Methods}}&    \multicolumn{3}{c}{CIFAR10} & \multicolumn{3}{c}{CIFAR100} & \multicolumn{3}{c}{ImageNet-100} \\
\cmidrule(rl){2-4}\cmidrule(rl){5-7}\cmidrule(rl){8-10}
                                 & All  & Old  & New  & All  & Old  & New  & All  & Old  & New \\
\midrule
$k$-means~\cite{macqueen1967some}  & 83.6 & 85.7 & 82.5 & 52.0 & 52.2 & 50.8 & 72.7 & 75.5 & {71.3} \\
RS+~\cite{han2021autonovel}          & 46.8 & 19.2 & 60.5 & 58.2 & {77.6} & 19.3 & 37.1 & 61.6 & 24.8 \\
UNO+~\cite{fini2021unified}               & 68.6 & \textbf{98.3} & 53.8 & 69.5 & 80.6 & 47.2 & 70.3 & \textbf{95.0} & 57.9 \\
ORCA~\cite{cao2021open}                     & 81.8 & 86.2 & 79.6 & 69.0 & 77.4 & 52.0 & 73.5 & {92.6} & 63.9 \\
\midrule
GCD~\cite{vaze2022generalized}       & {91.5} & {97.9} & {88.2} & {73.0} & 76.2 & {66.5} & {74.1} & 89.8 & 66.3 \\

CiPR~\cite{hao2023cipr}       & {97.7} & {97.5} & {97.7} & {81.5} & 82.4 & {79.7} & {80.5} & 84.9 & {78.3} \\
GPC~\cite{zhao2023learning}       & {90.6} & {97.6} & {87.0} & {75.4} & 84.6 & {60.1} & {75.3} & 93.4 & {66.7} \\
PromptCAL-1~\cite{zhang2023promptcal}       & {97.1} & {97.7} & {96.7} & {76.0} & 80.8 & {66.6} & {75.4} & 94.2 & {66.0} \\
PromptCAL-2~\cite{zhang2023promptcal}       & \textbf{97.9} & {96.6} & \textbf{98.5} & {81.2} & 84.2 & {75.3} & {83.1} & 92.7 & {78.3} \\
ComEx~\cite{yang2022divide} & 95.0 &92.6 &93.8 & 75.2&77.3 & 75.6 & - &-  &- \\
\midrule

SimGCD~\cite{wen2023simgcd}                    & {97.1} & {95.1} & {98.1} & {80.1} & {81.2} & {77.8} & {83.0} & {93.1} & {77.9} \\
\makecell[l]{SimGCD \\ +MCDL (Ours)}                    & {97.3} & {95.4} & {98.4} & \textbf{84.0} & \textbf{84.7} & \textbf{81.2} & \textbf{83.8} & {93.8} & \textbf{78.8} \\
$\Delta$                  & \textcolor{darkgreen}{\textbf{+0.2}} & \textcolor{darkgreen}{\textbf{+0.2}} & \textcolor{darkgreen}{\textbf{+0.3}} & \textcolor{darkgreen}{\textbf{+3.9}} & \textcolor{darkgreen}{\textbf{+3.5}} & \textcolor{darkgreen}{\textbf{+3.4}} & \textcolor{darkgreen}{\textbf{+0.8}} & \textcolor{darkgreen}{\textbf{+0.7}} & \textcolor{darkgreen}{\textbf{+0.9}} \\
\bottomrule
\end{tabular}
\end{center}
\caption{Results on \textbf{generic image recognition datasets}.} \label{subtab:generic}
\end{table}

\paragraph{Comparison with state-of-the-art methods.} In this section, we conducted a thorough experimental comparison with state-of-the-art methods in generalized category discovery. The methods we compared include SimGCD~\cite{wen2023simgcd}, ORCA~\cite{cao2021open}, GCD~\cite{vaze2022generalized}, GPC~\cite{zhao2023learning}, CiPR~\cite{hao2023cipr}, PromptCAL~\cite{zhang2023promptcal}, ComEx~\cite{yang2022divide}, as well as popular baselines derived from NCD, such as RS+~\cite{han2021autonovel} and UNO+~\cite{fini2021unified}. Additionally, we utilized the features from DINO for k-means clustering. Our method consistently outperforms both the baseline and previous methods in terms of overall accuracy across two generic image recognition datasets, as shown in Tab.~\ref{subtab:generic}. It also achieves competitive performance on the less challenging CIFAR-10 dataset. Although our results for old classes are inferior to UNO+ on the ImageNet-100 dataset, it is important to note that the goal of GCD is to discover new categories from the unlabeled set, thus highlighting the improvement in recognizing unseen categories as more crucial for generalized category discovery. Moreover, when compared with ComEx~\cite{yang2022divide}, a method that integrates the divide-and-conquer strategy, our MCDL introduce no extra networks for training and meanwhile significantly outperforms it with a large margin of 8.8\%/7.4\%/5.6\% for the accuracy of all/old/new classes respectively, demonstrating the superiority of our method. Furthermore, our MCDL demonstrates significant performance improvements on challenging semantic shift benchmarks by implicitly extracting valuable supervisory signals for unseen categories, as shown in Tab.\ref{subtab:ssb}, Notably, MCDL outperforms the baseline SimGCD by a substantial margin of 8.4\% and 8.1\% in terms of overall accuracy on the CUB and Stanford Cars datasets, respectively. It is worth highlighting that MCDL enhances the accuracy of both old and new class categories in all the results, demonstrating its generalization.

\begin{table}[t]
\begin{center}
\tablestylesmaller{3.2pt}{1.05}
\begin{tabular}{lccccccccc}
\toprule
{\multirow{2}{*}{Methods}}&   \multicolumn{3}{c}{CUB} & \multicolumn{3}{c}{Stanford Cars} & \multicolumn{3}{c}{FGVC-Aircraft}\\
\cmidrule(rl){2-4}\cmidrule(rl){5-7}\cmidrule(rl){8-10}
     & All  & Old  & New  & All  & Old  & New  & All  & Old  & New \\
\midrule
$k$-means~\cite{macqueen1967some}  & 34.3 & 38.9 & 32.1 & 12.8 & 10.6 & 13.8 & 16.0 & 14.4 & 16.8 \\
RS+~\cite{han2021autonovel}          & 33.3 & 51.6 & 24.2 & 28.3 & 61.8 & 12.1 & 26.9 & 36.4 & 22.2 \\
UNO+~\cite{fini2021unified}               & 35.1 & 49.0 & 28.1 & 35.5 & 70.5 & 18.6 & 40.3 & 56.4 & 32.2 \\
ORCA~\cite{cao2021open}                     & 35.3 & 45.6 & 30.2 & 23.5 & 50.1 & 10.7 & 22.0 & 31.8 & 17.1 \\
\midrule
GCD~\cite{vaze2022generalized}       & {51.3} & {56.6} & {48.7} & {39.0} & 57.6 & {29.9} & {45.0} & 41.1 & {46.9} \\
CiPR~\cite{hao2023cipr}       & {57.1} & {58.7} & {55.6} & {47.0} & 61.5 & {40.1} & {-} & - & {-} \\
GPC~\cite{zhao2023learning}       & {52.0} & {55.5} & {47.5} & {38.2} & 58.9 & {27.4} & {43.3} & 40.7 & {44.8} \\
PromptCAL-1~\cite{zhang2023promptcal}       & {51.1} & {55.4} & {48.9} & {42.6} & 62.8 & {32.9} & {44.5} & 44.6 & {44.5} \\
PromptCAL-2~\cite{zhang2023promptcal}       & {62.9} & {64.4} & {62.1} & {50.2} & 70.1 & {40.6} & {52.2} & 52.2 & {52.3} \\

\midrule
SimGCD~\cite{wen2023simgcd}                      & {60.3} & {65.6} & {57.7} & {53.8} & {71.9} & {45.0} & {54.2} & {59.1} & {51.8} \\
\makecell[l]{SimGCD \\ +MCDL (Ours)}                  & \textbf{68.7} & \textbf{74.2} & \textbf{63.4} & \textbf{61.9} & \textbf{77.9} & \textbf{54.7} & \textbf{57.6} & \textbf{64.6} & \textbf{52.5} \\
$\Delta$                    & \textcolor{darkgreen}{\textbf{+8.4}} & \textcolor{darkgreen}{\textbf{+8.6}} & \textcolor{darkgreen}{\textbf{+5.7}} & \textcolor{darkgreen}{\textbf{+8.1}} & \textcolor{darkgreen}{\textbf{+6.0}} & \textcolor{darkgreen}{\textbf{+9.7}} & \textcolor{darkgreen}{\textbf{+3.4}} & \textcolor{darkgreen}{\textbf{+5.5}} & \textcolor{darkgreen}{\textbf{+0.7}} \\
\bottomrule
\end{tabular}
\end{center}
\caption{Results on the \textbf{Semantic Shift Benchmark}~\cite{vaze2022generalized}. Significant performance boost can be achieved by our MCDL.} \label{subtab:ssb}
\end{table}

\begin{table}[t]
\begin{center}
\tablestyle{7.5pt}{1}
\resizebox{\linewidth}{!}{
\begin{tabular}{lcccccc}
\toprule
{\multirow{2}{*}{Methods}}&    \multicolumn{3}{c}{Vanilla ACC} & \multicolumn{3}{c}{Balanced ACC}\\
\cmidrule(r){2-4}
\cmidrule(r){5-7}
                                & All  & Old  & New  & All  & Old  & New \\
\midrule
$k$-means~\cite{macqueen1967some}  & 13.0 & 12.2 & 13.4 & 13.6 & 12.2 & 15.0 \\
RS+~\cite{han2021autonovel}          & 27.9 & {55.8} & 12.8 & - & - & - \\
UNO+~\cite{fini2021unified}               & 28.3 & {53.7} & 14.7 & - & - & - \\
ORCA~\cite{cao2021open}                     & 20.9 & 30.9 & 15.5 & 9.8 & 14.7 & 4.9 \\
\midrule
GCD~\cite{vaze2022generalized}       & {35.4} & 51.0 & {27.0} & 32.8 & 41.4 & 24.2 \\
CiPR~\cite{hao2023cipr}       & {36.8} & {45.4} & {32.6}  & {-} & - & {-} \\
PromptCAL~\cite{zhang2023promptcal}       & {37.0} & {52.0} & {28.9}  & {-} & - & {-} \\
\midrule
SimGCD~\cite{wen2023simgcd}   & {44.0} & {58.0} & {36.4} & {39.4} & {51.4} & {27.3} \\ 
\makecell[l]{SimGCD \\ +MCDL (Ours)} & \textbf{49.4} & \textbf{61.7} & \textbf{39.2} & \textbf{43.4} & \textbf{55.3} & \textbf{31.5} \\ 
$\Delta$                  & \textcolor{darkgreen}{\textbf{+5.4}} & \textcolor{darkgreen}{\textbf{+3.7}} & \textcolor{darkgreen}{\textbf{+2.8}} & \textcolor{darkgreen}{\textbf{+4.0}} & \textcolor{darkgreen}{\textbf{+3.9}} & \textcolor{darkgreen}{\textbf{+4.2}} \\
\bottomrule
\end{tabular}}
\end{center}
\caption{Results on more challenging \textbf{long-tailed fine-grained dataset} Herbarium19.}\label{tab:herb19}
\end{table}

Furthermore, in Tab.~\ref{tab:herb19}, we present the detailed results on the challenging long-tailed fine-grained dataset, Herbarium19, which closely resembles the real-world application of GCD. Given that the test split also follows a long-tailed distribution, it has the potential to obscure any biases present in the models. Therefore, following the approach in \cite{wen2023simgcd}, we report both the balanced accuracy (ACC) and the Vanilla ACC to provide a comprehensive evaluation of the effectiveness of our MCDL and evaluate other methods with the metrics as well. Our MCDL consistently achieves improvements on both metrics, showcasing its superiority over existing methods in real-world scenarios.

\begin{table}[t]
\begin{center}
\tablestyle{7.5pt}{1}
\resizebox{\linewidth}{!}{
\begin{tabular}{lcccccc}
\toprule
{\multirow{2}{*}{Methods}} & \multicolumn{3}{c}{CUB} & \multicolumn{3}{c}{CIFAR-100}\\
\cmidrule(r){2-4}
\cmidrule(r){5-7}
                    & All  & Old  & New  & All  & Old  & New \\
\midrule
GCD~\cite{vaze2022generalized}   & 51.3 & 56.6 & 48.7 & 73.0 & 76.2  &66.5   \\ 
GCD+MCDL & 55.6 & 63.1 &52.5  & 77.2 & 81.0  & 67.4  \\ 
$\Delta$ & \textcolor{darkgreen}{\textbf{+4.3 }} & \textcolor{darkgreen}{\textbf{+6.5 }} &  \textcolor{darkgreen}{\textbf{+3.8 }} & \textcolor{darkgreen}{\textbf{+4.2 }}  &  \textcolor{darkgreen}{\textbf{+4.8 }} & \textcolor{darkgreen}{\textbf{+0.9 }} \\ \midrule
CiPR~\cite{hao2023cipr}   &57.1  &58.7  & 55.6 & 81.5 &82.4   &79.7   \\ 
CiPR+MCDL &59.7  & 63.1 & 57.8 & 83.1 & 84.1  &80.5   \\ 
$\Delta$ & \textcolor{darkgreen}{\textbf{+2.6 }} & \textcolor{darkgreen}{\textbf{+4.4 }} &  \textcolor{darkgreen}{\textbf{+2.2 }} & \textcolor{darkgreen}{\textbf{+1.6 }}  &  \textcolor{darkgreen}{\textbf{+1.7 }} & \textcolor{darkgreen}{\textbf{+0.8 }} \\ 
\bottomrule
\end{tabular}}
\end{center}
\caption{\textbf{Comparison between popular GCD methods and their MCDL application} on CUB and CIFAR-100 datasets.}
\label{tab:generality}
\end{table}

\paragraph{Generality of MCDL.} To validate the generalization ability of our MCDL, we integrate it with two popular GCD methods, including GCD~\cite{vaze2022generalized}, CiPR~\cite{hao2023cipr}. As shown in Tab.~\ref{tab:generality}, MCDL brings consistent performance boost to all the baselines on both old and new categories across two datasets. For example, when seamlessly combined with MCDL, an accuracy increase of around 6.5\% on the old categories of CUB dataset can be achieved by GCD respectively, demonstrating that MCDL is capable of mining discriminative information from historical predictions. Therefore, all the results indicate the generalization ability of MCDL to boost existing methods.

\begin{table}[t]
\begin{center}
\tablestyle{7.5pt}{1}
\resizebox{\linewidth}{!}{
\begin{tabular}{lcccccc}
\toprule
{\multirow{2}{*}{Methods}} & \multicolumn{3}{c}{CUB} & \multicolumn{3}{c}{CIFAR-100}\\
\cmidrule(r){2-4}
\cmidrule(r){5-7}
                    & All  & Old  & New  & All  & Old  & New \\
\midrule

Baseline   & {60.3} & {65.6} & {57.7} & 80.1 & 81.2 & 77.8 \\
\midrule
+ $\mathcal{M}_w$ & 66.8 & 72.9 & 60.7 & 82.7 &83.7  &80.1  \\ 
+ $\mathcal{M}_s$ & 62.8 & 65.0 & 58.9 & 81.5 &82.4  &78.5 \\  
+ $\mathcal{M}_w$\&$\mathcal{M}_s$ &\textbf{68.7} & \textbf{74.2} & \textbf{63.4} & \textbf{84.0} & \textbf{84.7}  & \textbf{81.2} \\ \midrule
+ $\mathcal{L}_{{sup}}$& 67.2 & 73.6 &61.1  & 82.6 &83.5  & 80.2 \\
+ $\mathcal{L}_{{sup}} \& \mathcal{L}_{{semi}}$& 68.1 & 73.9 &62.4  & 83.4 & 84.2 & 80.9 \\
+ $\mathcal{L}_{{sup}} \& \mathcal{L}_{{semi}} \& \mathcal{L}_{{self}}$& \textbf{68.7} & \textbf{74.2} & \textbf{63.4} & \textbf{84.0} & \textbf{84.7}  & \textbf{81.2} \\
\bottomrule
\end{tabular}}
\end{center}
\caption{\textbf{Ablation study for the effectiveness of the proposed DCM and DCL} on CUB and CIFAR-100 datasets.}
\label{tab:ablation}
\end{table}

\subsection{Ablation studies}
\paragraph{Component analysis.} To study the influence of each component within the proposed MCDL, we conduct detailed ablation analysis on CUB and CIFAR-100 datasets. For the DCM module, as shown in Tab.~\ref{tab:ablation}, when only $\mathcal{M}_w$ is used for credibility modelling, both the performance of old and new classes can be improved significantly, especially on the challenging CUB dataset. Specifically, on the CUB dataset, $\mathcal{M}_w$ can improve the baseline by a large margin of 6.5\%/7.3\%/3.0\% for all/old/new classes respectively. However, when only adopting $\mathcal{M}_s$, it achieves inferior performance to $\mathcal{M}_w$. This is mainly due to the fact that the predictions suffer from severe fluctuations caused by the disturbance of strong augmentation, resulting in unreliable pseudo labels and noisy sample selections. Additionally, the use of both memory banks can further boost performance by considering both intra-memory consistency and inter-memory consistency and thus achieve the best results. Furthermore, in the DCL module, $\mathcal{L}_{sup}$ fully utilizes the discriminative information within the high-quality one-hot labels generated by highly consistent historical predictions, thus improving the baseline by 6.9\%/2.5\% on two datasets. When combined with $\mathcal{L}_{semi}$, a performance boost from 67.2\% to 68.1\% on the CUB dataset can also be observed. This verifies the presence of valuable supervisory signals in the predictions of lower consistency, and the semi-supervised learning process can effectively extract discriminative information from the noisy labels and meanwhile alleviate the negative influence of label noise. Finally, $\mathcal{L}_{self}$ contributes to the best performance as well by enhancing prediction consistency and providing noise-free self-supervised signals for training.


\begin{table}[t]
\begin{center}
\tablestyle{7.5pt}{1}
\resizebox{0.8\linewidth}{!}{
\begin{tabular}{lccc}
\toprule
{{Methods}} & {CUB} & {CIFAR-100} & {CIFAR-10}\\

\midrule
$\mathcal{M}_w$ &70.58  & 72.24 & 80.76   \\ 
$\mathcal{M}_s$ &64.29 & 68.12 & 73.32 \\  
$\mathcal{M}_w$\&$\mathcal{M}_s$ &\textbf{72.92} & \textbf{76.45} & \textbf{84.71}  \\
\bottomrule
\end{tabular}}
\end{center}
\caption{\textbf{Ablation study for the influence of $\mathcal{M}_w$ and $\mathcal{M}_s$} on the label accuracy of the selected high-credibility samples.}
\label{tab:labelaccuracy}
\end{table}

\begin{figure*}[!t]
    \centering
    \includegraphics[width=1.0\linewidth]{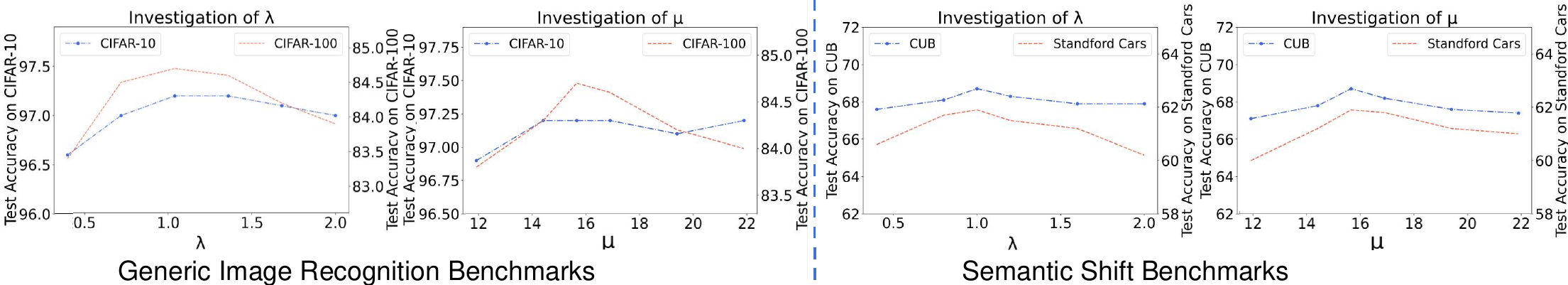}
    \caption{\textbf{Investigation of $\lambda$ and memory bank length $\mu$} on both generic image recognition (i.e., CIFAR-10/100) and semantic shift benchmarks (i.e., Standford Cars, CUB). Similar performance can be achieved across different values for both two parameters.
    }
    \label{fig:ablationstudy}
\end{figure*}

\begin{figure*}
        \centering
    \includegraphics[width=1.0\linewidth]{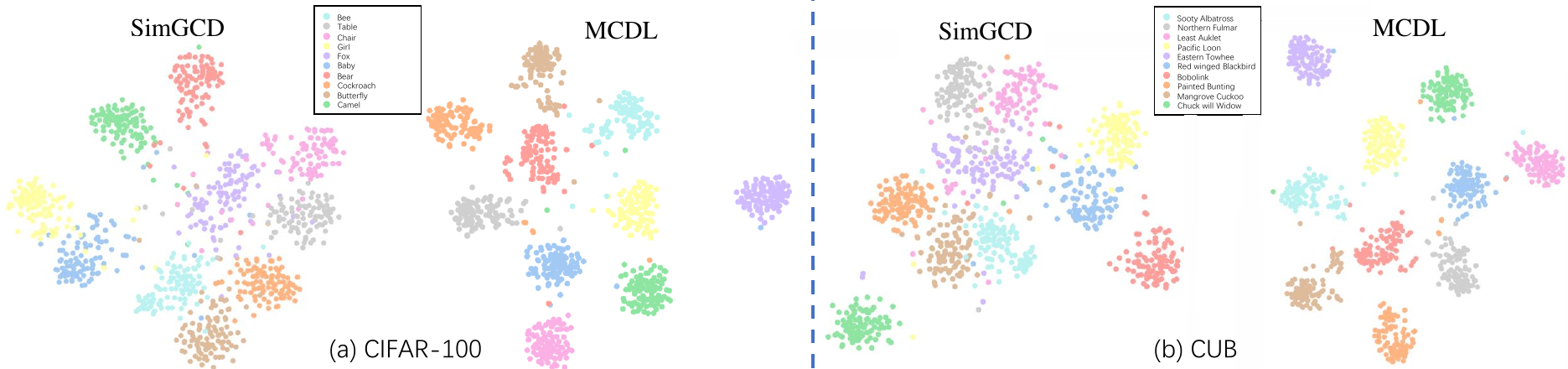}
    \caption{\textbf{The qualitative  comparison between SimGCD and our SimGCD+MCDL.} We plot the t-SNE visualization of the representations of unlabeled samples on CIFAR-100 and CUB datasets. The samples are randomly m 10 unseen classes.
    }
    \label{fig:tsne}
\end{figure*}

\paragraph{Analysis on memory banks.} Here we conduct detailed investigation on the effect of two memory banks. As shown in Tab.~\ref{tab:labelaccuracy}, when only utilizing for credibility modeling, the label accuracy of the selected training samples with high-credibility decreases. This decline can be primarily attributed to the fact that weak augmentation has minimal impact on the sample prediction. Furthermore, the presence of confirmation bias leads to similar historical predictions stored in the memory bank. Consequently, the pseudo labels generated using such memory bank are unreliable and noisy. And the utilization of such noisy labels for supervised training further degrades the quality of representation. As a result, the reliability of sample prediction is reduced, leading to a decline in the label accuracy of the select samples. Similarly, when we only use the strongly-augmented memory bank for sample selection, the consistency of historical predictions is poor due to the large fluctuation of the predictions, resulting in unreliable sample selection as well. However, when we take both weakly-augmented and strong-augmented memory banks into consideration, the results in both Fig.~\ref{fig:accuracy} and Tab.~\ref{tab:labelaccuracy} demonstrate that the label accuracy is improving across the training process. This is mainly account for that the shortcomings of each memory bank can compensate for the other, which enables more accurate selection of correct samples from the unlabeled set and alleviates over-fitting to noisy labels meanwhile.

\paragraph{Parameter sensitivity.} Furthermore, we investigate how the memory bank length $\mu$ and the balanced loss weight $\lambda$ affect the performance. Fig.~\ref{fig:ablationstudy} shows results on both generic recognition and semantic shift benchmarks. It can be observed that similar performance can be achieved by MCDL across different values of two parameters on both two datasets. Thus our MCDL is not sensitive to the value of both $\lambda$ and $\mu$. Specifically, increasing the length of memory bank helps achieve slightly higher accuracy at first and the performance remains nearly the same when $\mu$ is larger than 16 since the information in the memory bank is sufficient for accurate sample selection and thus brings no further boost. For the loss weight, similar phenomenon can be observed at first as well, but when further increasing the weight, there is a slight drop in the accuracy since it may break the trade-off between new and old classes and thus impair the performance of old classes. Therefore, we set $\lambda$ and $\mu$ as 1.0 and 16 in our experiments.

\paragraph{Qualitative results.} As shown in Fig.~\ref{fig:tsne}, we use t-SNE to visualize the distribution of the features of SimGCD and SimGCD+MCDL for samples from 10 randomly-selected unseen categories in both CUB and CIFAR-100 datasets. This visualization helps us gain insight into the performance improvement achieved by our MCDL. The results show that MCDL greatly enhances the quality of features for the unseen class, even on the challenging semantic shift benchmark. Such results further verify the effectiveness of our MCDL in extracting valuable discriminative information from historical predictions to establish reliable supervisory signals for learning the unseen class.

\section{Conclusion}
In this paper, we propose a Memory Consistency guided Divide-and-conquer Learning framework (MCDL) for Generalized Category Discovery. Specifically, we empirically
discover that the predicted categories of samples with consistent historical predictions are more likely to align with the ground truth labels. Inspired by this, we propose a dual-consistency credibility modeling strategy for each sample by taking both intra-memory and inter-memory consistency into consideration. Furthermore,  we design a divide-and-conquer learning framework to enhance the performance with discriminative information existed in samples with relatively high credibility and meanwhile tackle with the influence of label noise. We demonstrate consistent performance improvements over previous methods on four generic classification datasets and four challenging semantic shift benchmarks. Our method also exhibits a good generality, which can be integrated with existing methods as a plug-in-play strategy to obtain further improvement.

\medskip

{\small
\bibliographystyle{ieee_fullname}
\bibliography{egbib}
}

\end{document}